\documentclass{bmvc2k}
\usepackage{amsmath}
\usepackage{bm}
\usepackage{amssymb}
\usepackage{booktabs}
\usepackage{float}
\usepackage{array}
\usepackage{pdfpages}

\usepackage{stringstrings} 

\newcommand{\change}[1]{#1} 

\title{\MODEL{}: An Unsupervised Implicit 3D Model of
Articulated Human Feet}

\addauthor{Oliver Boyne}{ob312@cam.ac.uk}{1}
\addauthor{James Charles}{http://www.jjcvision.com}{1}
\addauthor{Roberto Cipolla}{cipolla@eng.cam.ac.uk}{1}

\addinstitution{
 Machine Intelligence Lab \\
 Department of Engineering \\
 University of Cambridge \\
 Cambridge, U.K.
}

\runninghead{Boyne, Charles, Cipolla}{\MODEL{} - 3D Model of Articulated Human Feet}

\def\eg{\emph{e.g}\bmvaOneDot}

\def\etal{\emph{et al}\bmvaOneDot}

\newcommand{\inspace}[1]{\in \mathbb{R}^{#1}}

\def\MODEL{FIND}
\def\partloss{unsupervised part-based loss}

\begin{document}

\maketitle

\begin{abstract}

In this paper we present a high fidelity and articulated 3D human foot model. The model is parameterised by a disentangled latent code in terms of shape, texture and articulated pose. While high fidelity models are typically created with strong supervision such as 3D keypoint correspondences or pre-registration, we focus on the difficult case of little to no annotation. To this end, we make the following contributions: (i) we develop a \textbf{F}oot \textbf{I}mplicit \textbf{N}eural \textbf{D}eformation field model, named \MODEL{}, capable of tailoring explicit meshes at any resolution~\textit{i.e.} for low or high powered devices; (ii) an approach for training our model in various modes of weak supervision with progressively better disentanglement as more labels, such as pose categories, are provided; (iii) a novel \partloss{} for fitting our model to 2D images which is better than traditional photometric or silhouette losses; (iv) finally, we release a new dataset of high resolution 3D human foot scans, \textit{Foot3D}. On this dataset, we show our model outperforms a strong PCA implementation trained on the same data in terms of shape quality and part correspondences, and that our novel \partloss{} improves inference on images.

\end{abstract}

\section{Introduction}
\label{sec:intro}

Shape reconstruction from single-view or few-view images is a vital but difficult computer vision task. Current approaches are dependent on strong priors to help constrain this ill-posed problem, usually in the form of well constructed and parameterised 3D models~\cite{loper2015smpl,zuffi20173d}. The process of model creation is also often time consuming, requiring a lot of supervision e.g. registration and correspondence annotation. For the case of human feet, the current state-of-the-art in reconstruction involves the use of expensive scanning equipment \cite{Volumental,yeti3d,ArtecLeo},  unavailable to the average consumer at home. Thus, there is growing interest for easier methods of foot reconstruction, particularly for home health monitoring, custom orthotics and the growing online shoe retail industry. As such, these models must be capable of running on low powered devices such as mobile phones, but also displayable at high resolution ~\textit{e.g.} for a doctor's inspection. To this end, we develop \MODEL{}, a \textbf{F}oot \textbf{I}mplicit \textbf{N}eural \textbf{D}eformation field model, which is easy to train as it requires little to no labels and can produce explicit meshes at tailored resolutions, useful for targeting specific hardware or needs.  \change{We can improve the model further by introducing minimal extra supervision: (i) we add the weak label of foot identities to disentangle shape and pose; and (ii) pose descriptions to produce a more interpretable pose latent space.}\\

We outline our key contributions as follows: (1) a novel implicit foot model, \MODEL, capable of producing high fidelity human feet and fully paramaterised by a disentangled latent space of shape, texture and pose. The model can also be tuned to specific hardware considerations or needs; (2) an approach to train the model under various levels of weak supervision, with stronger supervision producing better disentanglement; (3) a novel, \partloss{} for fitting \MODEL{} to images using unsupervised feature learning; and (4) we release \textit{Foot3D}, a dataset of high resolution foot scans, providing shape and texture information for training of 3D foot models.


\begin{figure}
   \includegraphics[width=\textwidth]{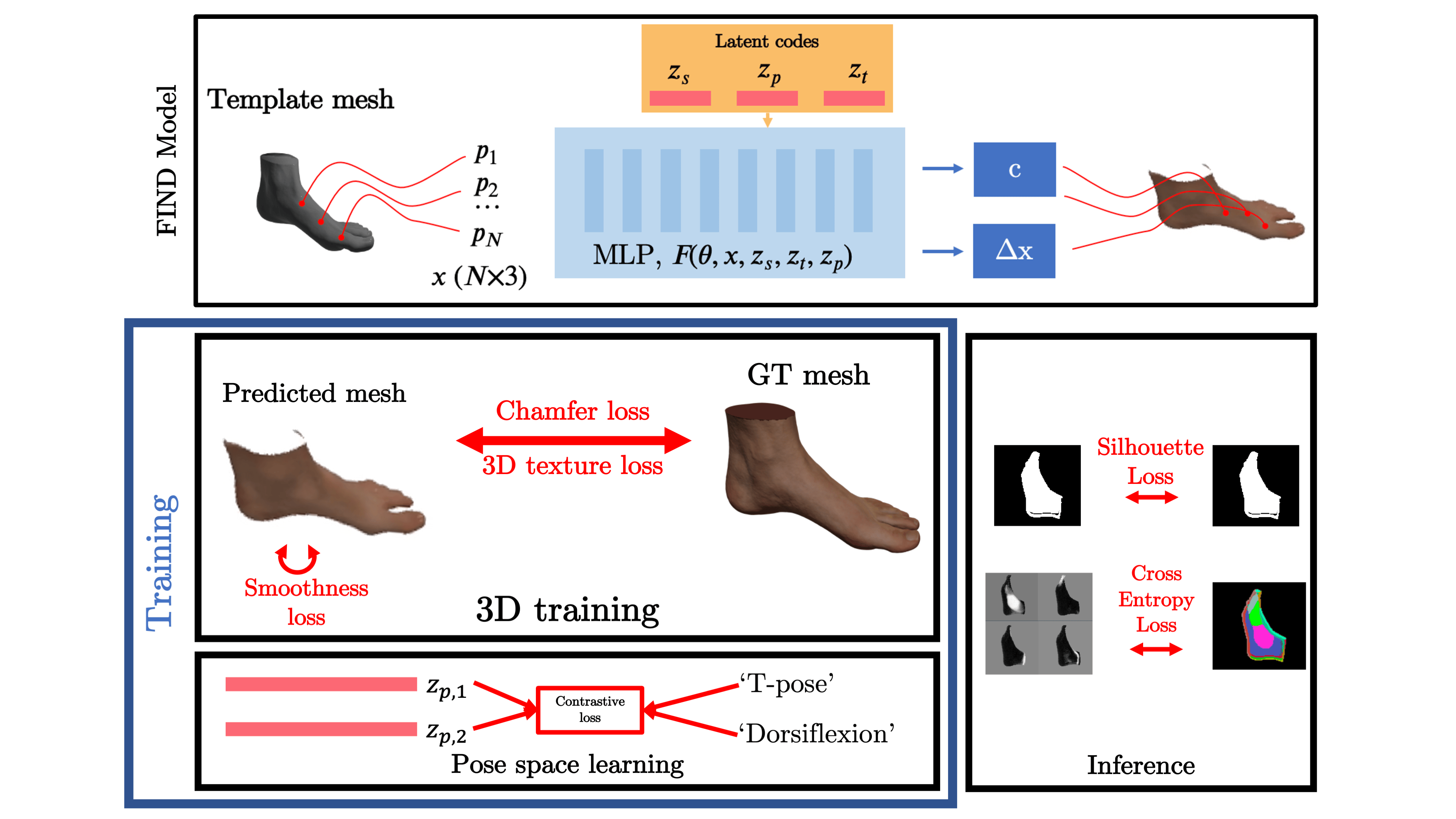}
   \vspace{-15pt}
   \caption{We produce (1) a coordinate based multilayer perceptron (MLP) architecture, \MODEL, which receives a 3D vertex position on a template mesh, and a shape, pose and texture embedding, and predicts a corresponding vertex deformation and colour. (2) We train our model with 3D based losses against ground truth scans, and learn a sensible pose space using a contrastive loss (3) We use differentiable rendering at inference to enforce a silhouette based loss, in addition to our novel \partloss{}. }
   \vspace{-5mm}
\end{figure}

\section{Related work}

\paragraph{Generative shape models.} These models endeavour to capture the distribution of an object category's shape in a set of controllable parameters~\textit {e.g.} height, width, pose, etc. Much of the work in generative shape modelling looks at human bodies where a number of parameterised models have been built~\cite{loper2015smpl, zuffi2015stitched}. For construction, the models are typically trained with strong supervision via 2D annotation~\cite{andriluka14cvpr, lin2014microsoft} of rich 3D scans~\cite{h36m_pami, vonMarcard2018}. The building of generative models of more niche object categories often uses similar principles, but usually on much smaller, custom datasets, often due to the time and expense in collecting the data. For examples of the difficulties involved, analogous generative quadruped models have been constructed from scans of toys of real animals~\cite{zuffi20173d} (as real animals don't keep still long enough), and parametric hand models have been constructed from 3D scans~\cite{romero2022embodied} and even at larger expense from MRI scans~\cite{li2021piano}. These models typically use Principal Component Analysis (PCA) to compose a small set of parameters which control linear offsets from a template mesh. We show here, that our implicit surface deformation field performs better against this type of strong PCA baseline on feet.




\paragraph{Implicit shape representations.} Recent interest has shifted towards implicit representations of shape - functions, often deep networks, that take in coordinates corresponding to a point in space, and return spacial information relating to an object - \change{such as } whether the point in space is inside the object \cite{chen2019learning}, or how near that point is to the object's surface \cite{park2019deepsdf}. These methods are naturally differentiable and can represent scenes at arbitrary scales, so lend themselves to many reconstruction tasks \change{ - for example, reconstructing meshes from point clouds \cite{hanocka2020point2mesh} or single view images \cite{wang2018pixel2mesh}}.
 One particular method, Neural Radiance Fields (NeRFs)~\cite{mildenhall2020nerf}, has become increasingly popular due to the very high fidelity in which it can synthesis novel views of a scene. Many extensions to NeRF have investigated how to parameterise shape and texture for generic object classes~\cite{jang2021codenerf}, decompose scenes into individual, controllable components~\cite{niemeyer2021giraffe}\change{, and directly make edits to geometry and texture \cite{liu2021editing}}. This has also been extended to posed humans, with several works~\cite{su2021nerf, xu2021h} showing promise in modelling dynamic human models using NeRF architectures. We take inspiration from this approach and, similar to~\cite{michel2022text2mesh}, build our implicit foot model \MODEL{} by sampling points on the surface of a manifold, which in our case are vertices of a high resolution template foot mesh, and learn a parameterised deformation field for deforming vertices to different types of foot shapes and poses.


\paragraph{Unsupervised representation learning for 3D objects.}
Generative Adversarial Networks (GANs)~\cite{goodfellow2014gan} have shown promise in learning unsupervised representations of 2D data, even learning information about the underlying 3D geometry~\cite{pan20202d}. The intermediate feature representations are sufficiently powerful that Zhang~\etal~\cite{zhang2021datasetgan} showed they can be trained to solve downstream tasks, such as semantic segmentations, with very simple MLP classifiers and an incredibly small amount of ground truth labels. We leverage this for our novel \partloss, allowing us to learn foot part classes in 3D (from 2D images) to aid fitting to 2D images.


\paragraph{Foot reconstruction.} Solutions exist for high resolution, accurate 3D scanners~\cite{Volumental,yeti3d,ArtecLeo}. These are expensive and difficult to use outside of specific environments, so do not lend themselves to the average consumer.
While PCA parameterised foot models do exist~\cite{amstutz2008pca}, many existing foot reconstruction solutions instead directly predict an unconstrained mesh from pointclouds or depth maps~\cite{lunscher2017point}.
Large scale, proprietary datasets of feet, scanned at low resolution~\cite{jurca2019analysis, jurca2011dorothy} only release the population measurement statistics to the public. For our work, we build a new foot scan dataset, \textit{Foot3D}, for model building and evaluation, which we release to the community.
\vspace{-12pt}
\section{Method}
\vspace{-5pt}
\subsection{Foot3D Dataset}

\newcommand{\gtfootimg}[1]{\includegraphics[width=0.175\textwidth]{images/dataset/images-29-07/#1}}

\newcolumntype{P}[1]{>{\centering\arraybackslash}p{#1}}
\begin{figure}
   \centering
   \small
   \begin{tabular}{*{5}{P{0.17\textwidth}}}
   \gtfootimg{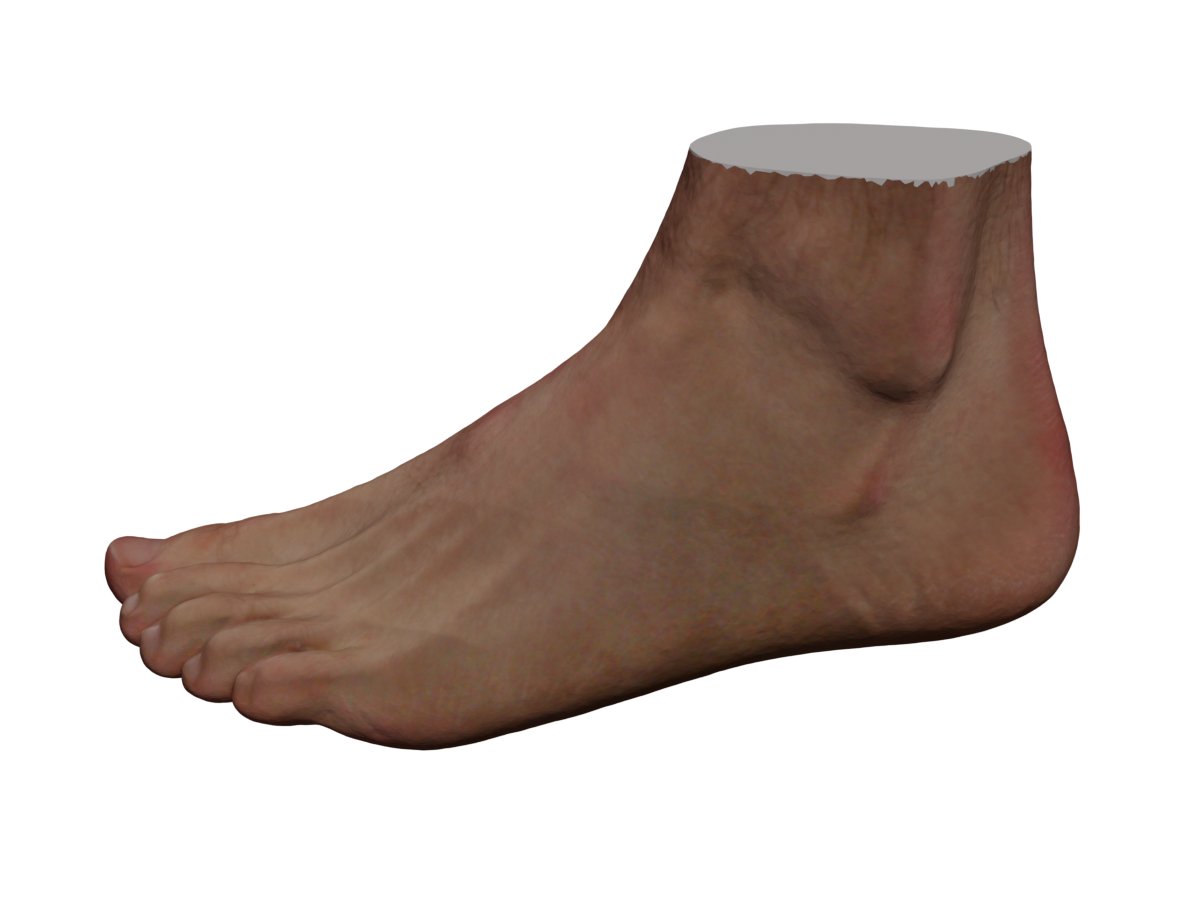} & \gtfootimg{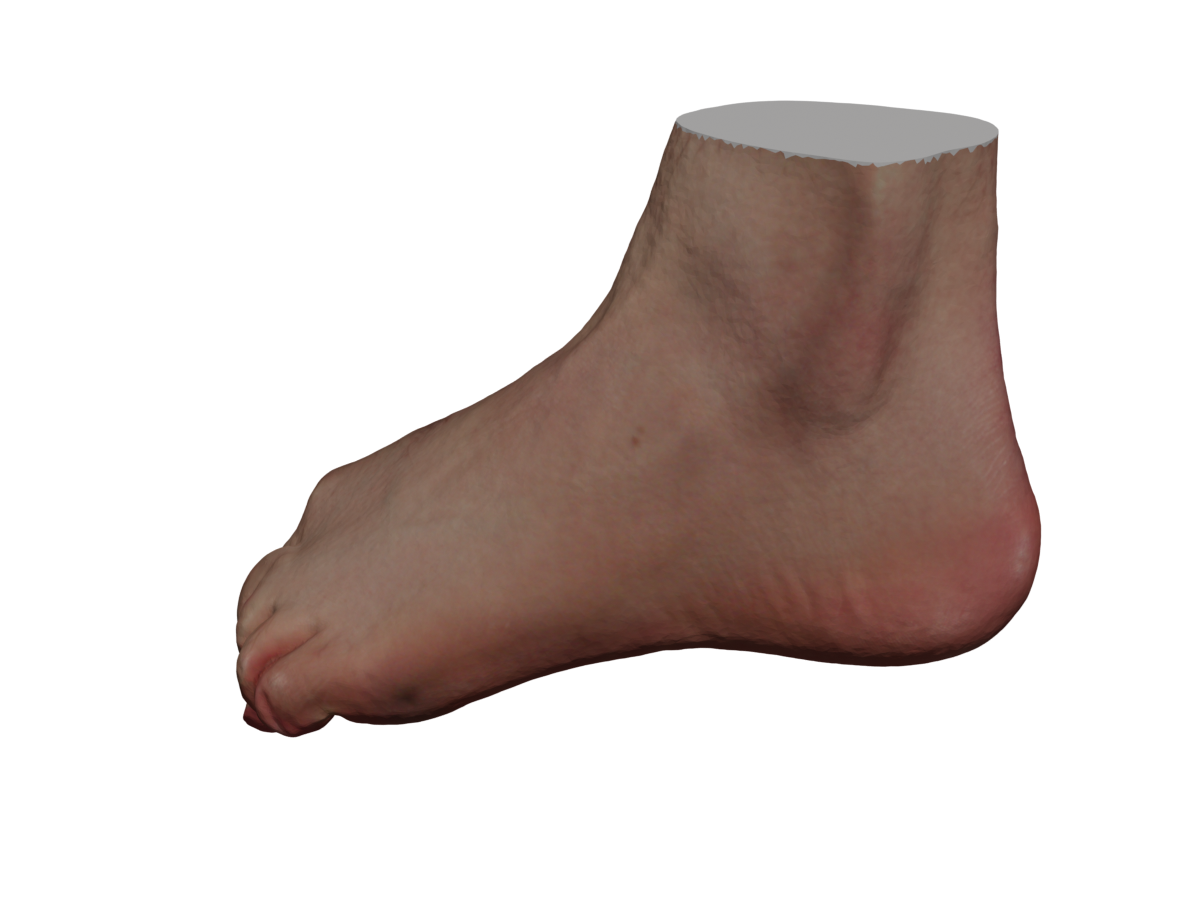} & \gtfootimg{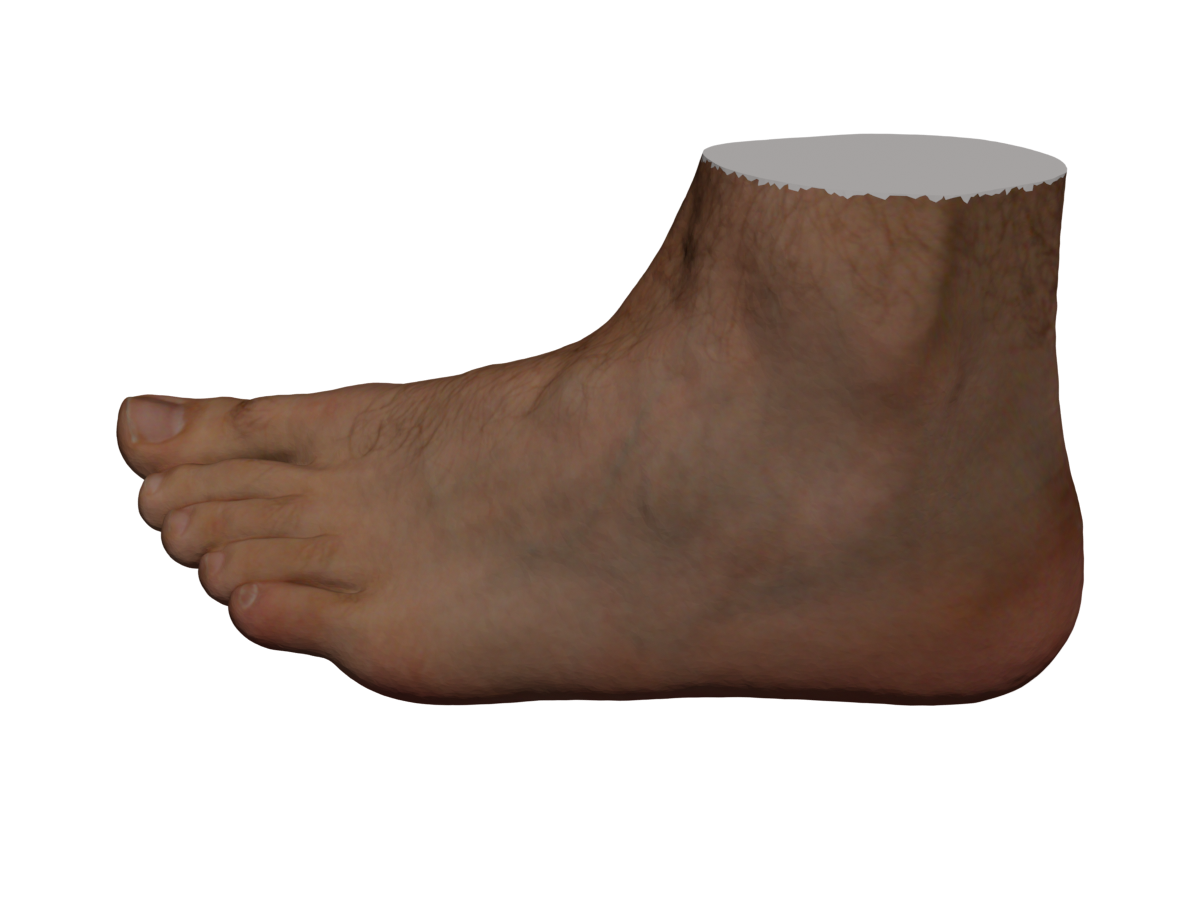} & \gtfootimg{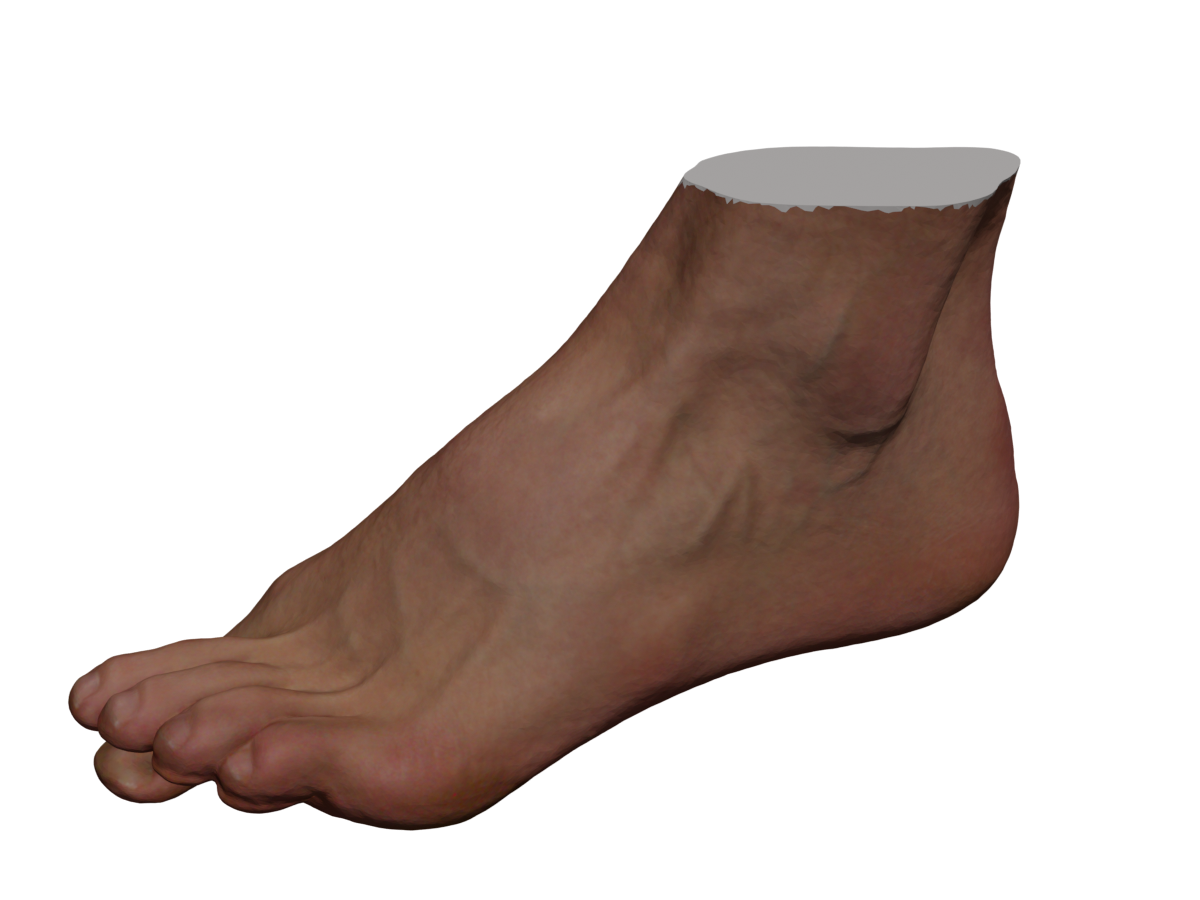} & \gtfootimg{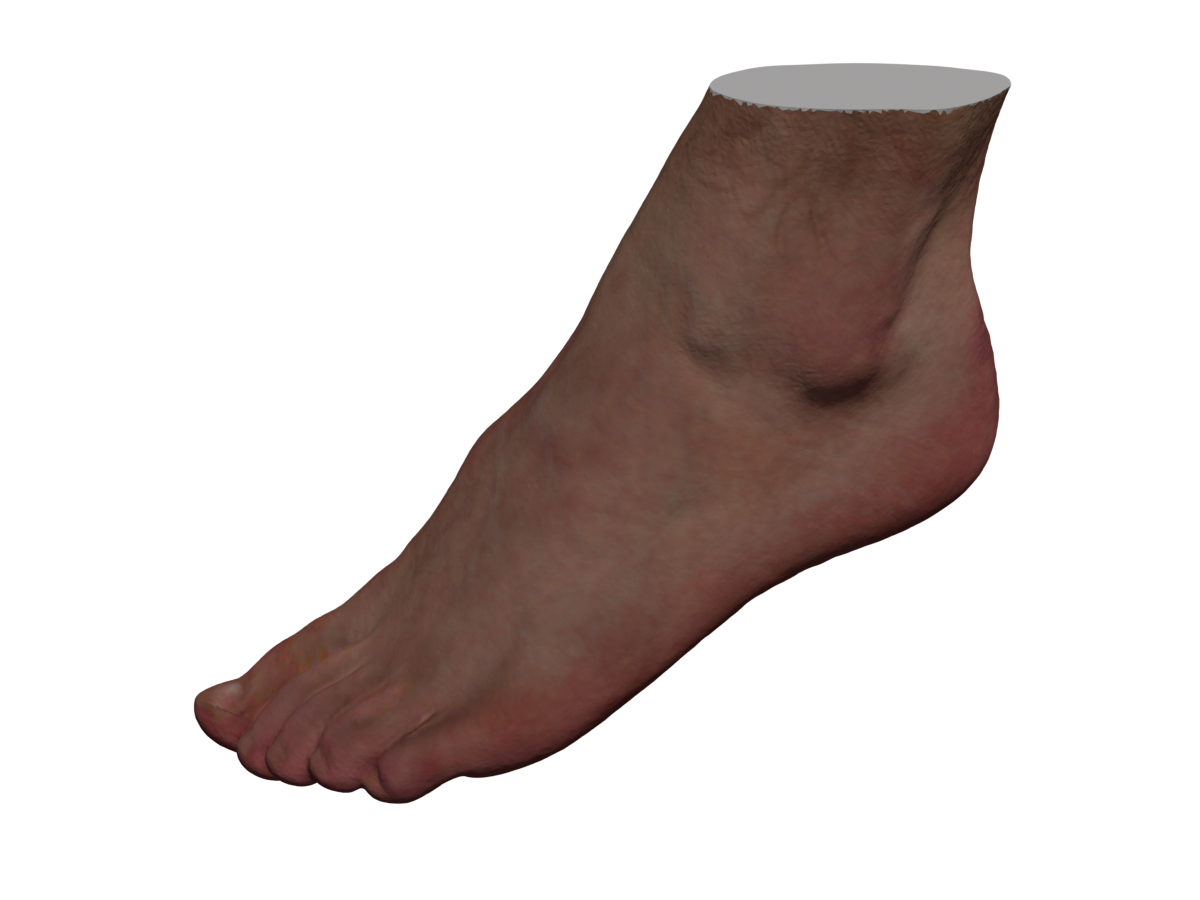}\\
    T-Pose & Toe flexion & Medial rotation, Dorsiflexion & Toe Abduction & Eversion\\
   \end{tabular}
   \vspace{-10pt}
   \centering
   \label{fig:dataset}
   \caption{5 feet from our dataset and corresponding pose descriptions. \change{The pose types are based on foot articulation described in foot anatomy literature \cite{houglum2011brunnstrom} - further details in the supplementary}}
\end{figure}

\def \ntpose {27}
\def \narticulated {51}
\def \ntotal {118}
\def \lefttotal {61}

We produce \textit{Foot3D}, a dataset of high resolution, textured 3D human feet. For acquisition, we use an Artec Leo 3D scanner \cite{ArtecLeo}, which has a 3D point accuracy of up to 0.1mm. A total of \lefttotal{} scans of the left feet on 34 subjects in a variety of poses was collect. To capture the entire surface of the foot, subjects would sit with their leg on a table for stability, and their foot suspended over the edge, this allowed the scanner to view the entire foot surface. Subjects then hold a static pose for approximately 2 minutes while the scan takes place. Details of the nature of the articulation requested can be found in the supplementary.
\vspace{-5mm}

\paragraph{} The raw data is then processed in Artec Studio 16 \cite{ArtecStudio} to produce 100K polygon meshes. These meshes were cut-off at an approximate position on the shin, by a plane approximately perpendicular to the leg.
Next, we leverage this slice-plane as a basis for loosely registering all of the meshes in the dataset, such that these planes are parallel to the XY-plane and a vector from the slice plane to the foot's centroid lies in the X direction. We then slice each foot at a uniform height above the heel to provide a consistent ankle length. Figure \ref{fig:dataset} shows dataset samples, and further details about this process are provided in the supplementary.

\vspace{-6pt}



\subsection{The \MODEL{} model}

\begin{figure}
\centering
\includegraphics[width=\textwidth]{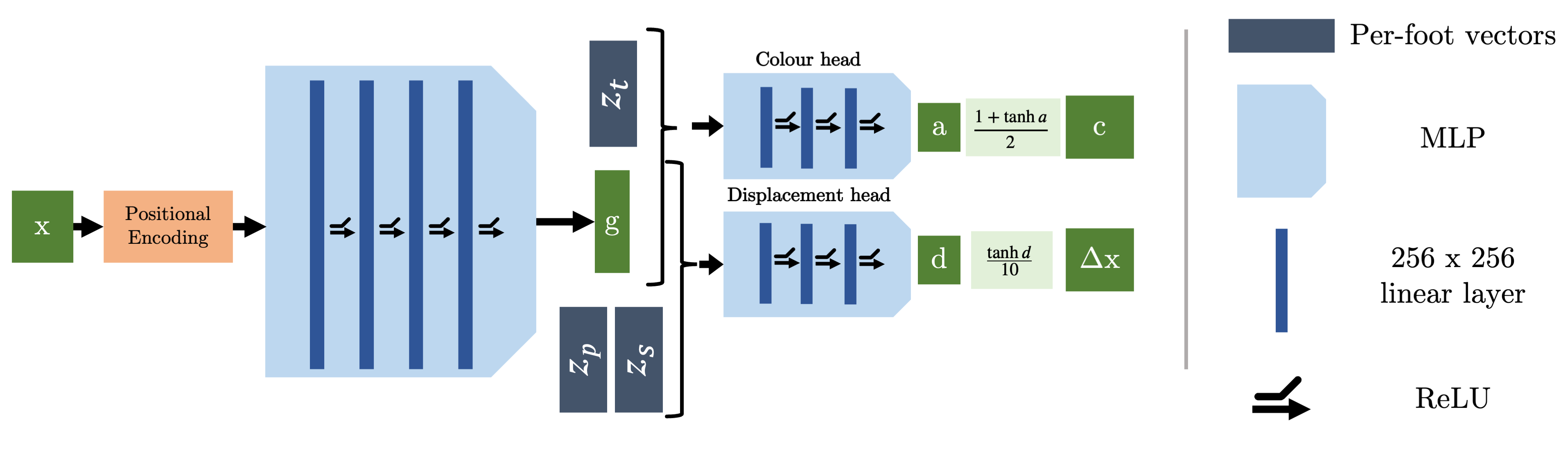}
\caption{Our implicit surface field is an MLP which takes as input a 3D point queried on a template mesh's surface, and texture, shape and pose embeddings, and provides as output a vertex colour value and displacement.}
\label{fig:model-overview}
\end{figure}

\def\offsets{\Delta x}
\def\templverts{v_\textrm{t}}
\def\sampleverts{x}
\def\shapecode{z_s}
\def\texcode{z_t}
\def\posecode{z_p}

The \MODEL{} model is controlled using a shape, pose and texture embedding to produce a surface deformation and colour field over a template mesh. This function is implemented as an implicit coordinate based neural network. The architecture is now explained in detail.
\paragraph{Positional Encoding. } Texture and shape have both high frequency and low frequency information across a foot's surface. As a result, we leverage positional encoding to allow for the network to capture the higher frequency signals \cite{mildenhall2020nerf, rahaman2019spectral},
\begin{equation}
\gamma_N(x) = \{x, \sin(2\pi x), \cos(2 \pi x), \dots, \sin(2N\pi x), \cos(2N\pi x)\}
\end{equation}
\vspace{-10mm}

\paragraph{Network. } After we encode our position with $N=10$, we learn a network $F$ with weights $\theta$, which takes shape, pose and texture encodings $\shapecode$, $\posecode$ and $\texcode$, and predicts vertex offsets $\offsets$ relative to a template mesh and vertex colours $c$,

\begin{equation}
F(\theta, \gamma(x), \shapecode, \posecode, \texcode) \rightarrow (\offsets, c)
\end{equation}

The main body of the network is 4 $256 \times 256$ linear layers, encoding the  \change{positional encoding $\gamma(x)$} into a feature, $g$.
\vspace{-2mm}
\paragraph{Heads.} We concatenate $g$ with the relevant shape codes and pass to a head specific for each task. The colour head receives $g$ and the texture code $\texcode$, and the displacement head receives $g$ and the shape and pose codes, $\shapecode$ and $\posecode$.
Both heads are composed of 3 $256 \times 256$ linear layers. We choose $\tanh$ as our activation function, and normalize outputs such that colour values are in the range $[0, 1]$, and displacements in the range $[-0.1, 0.1]$.

\vspace{-2mm}
\paragraph{Defining the surface.} We define the surface over which the network is trained by using a template mesh, which we take as a foot in our dataset outside of our training and validation sets. 
The vertex positions $\sampleverts$ are found by sampling across the surface of a template mesh. At inference time, all of these vertices can be used to output a full mesh.
\vspace{-2mm}
\paragraph{Latent vectors.} The shape, texture and pose encodings, all of size 100, are jointly learned during training. We place the constraint that texture and shape codes must be shared between different scans of the same foot. We also learn a per-foot registration to the ground truth: an Euler rotation $r \inspace{3}$, global translation $t \inspace{3}$, and global XYZ scaling $s \inspace{3}$.

\begin{figure}
   \centering
   \includegraphics[width=0.8\textwidth]{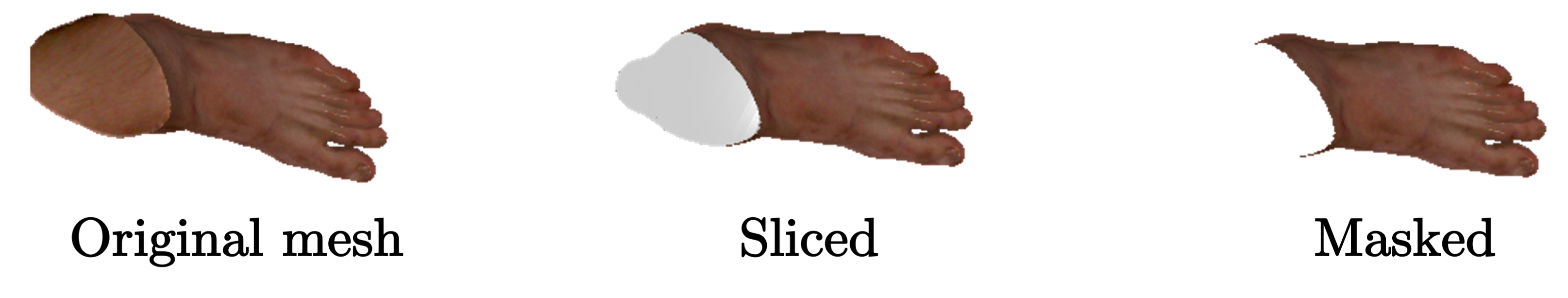}
   \caption{We show the effect of slicing and masking on our dataset and our differentiable renderer.}
   \label{fig:slice-render}
\end{figure}

\subsection{Learning part-segmentation}

Here we aim to develop an inference-time supervisory loss that learns part-awareness in an unsupervised manner.
\vspace{-3mm}
\paragraph{Differentiable rendering.} In order to optimise the model to fit to real images, it is necessary to be able to differentiably render the model. We use PyTorch3D’s \cite{ravi2020pytorch3d} differentiable renderer. As feet are not complete, watertight objects on their own, we choose to mask out the slice plane applied to the feet models when rendering, as we argue this better reflects how real feet appear in images (post-segmentation). To achieve this, all faces on this slice plane are marked when formulating the dataset, and these faces are masked out of the renders produced by the differentiable renderer. Figure \ref{fig:slice-render} shows the effect of this process.
\vspace{-2mm}

\paragraph{Extracting features. } Using the knowledge that 2D GANs can learn 3D geometrical information~\cite{pan20202d}, we train a StyleGAN2~\cite{karras2020analyzing} network to generate rendered feet from StyleGAN2 latent codes. We train on 100k 128x128 rendered images from models within Foot3D, with camera viewpoints in a hemisphere above the foot.
To use this network as an inference loss, invert StyleGAN to obtain the codes using the Restyle~\cite{alaluf2021restyle} encoder.
When passing a rendered image from our differentiable renderer through the encoder-decoder, similar to~\cite{zhang2021datasetgan}, we can use the features maps after the AdaIN layers of StyleGAN2 as a form of high level information about the foot in the image.   

\begin{figure}
   \centering
   \includegraphics[width=1.0\textwidth]{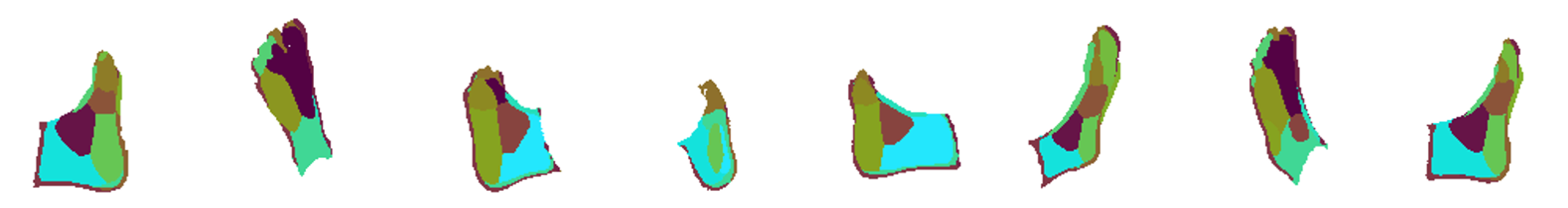}
   \vspace{-10pt}
   \caption{20 clusters on the 8th feature map of our Restyle encoder.}
   \label{fig:restyle-cluster}
   \vspace{-4mm}
\end{figure}

\vspace{-2mm}
\paragraph{Learning unsupervised 2D parts. } To learn foot parts from these feature maps, we observe that clustering them with k-means produces meaningful part segmentations, as can be seen in Figure \ref{fig:restyle-cluster}, which correspond between different feet. As a result, we train a simple linear classifier to produce these per-pixel part segmentations  from a subset of the StyleGAN feature maps. The whole classification process is fully differentiable.

\vspace{-2mm}
\paragraph{Moving to 3D.} We then use this classifier to learn a per-vertex part probability on the template mesh~\textit{i.e.} a $C$ length vector $c \in [0,1]$ where $C$ is the total number of classes. These vectors can be directly `rendered' using our differentiable renderer to any new $C$ channel image producing a per pixel foot part probability map.

\vspace{-2mm}
\paragraph{Unsupervised part-based loss.} Using this part classification pipeline, we enforce a cross-entropy loss on our $C$ cluster classes when optimising to images. Given our rendered part probability map $P \inspace{H\times W \times C}$, and our `ground-truth' labels from our classifier $L \inspace{H\times W}$, we define our loss,
\vspace{-2mm}
\begin{equation}
Loss_{\textrm{CE}} = - \frac{1}{H \times W} \sum^H_{y=1}{\sum^W_{x=1}{\sum^C_{c=1} \ln{\frac{\exp{P_{y,x,L_{y,x}}}}{\sum^C_{i=1}\exp{P_{y, x,i}} } } } }  
\end{equation}

\vspace{-5mm}
\subsection{Training Losses}

Our training loop and inference process uses several different losses:
\vspace{-2mm}

\newcommand{\norm}[1]{\left\lVert#1\right\rVert}
\def \chamf {L_\textrm{chamf}}
\def \smooth {L_\textrm{smooth}}
\def \tex {L_\textrm{tex}}
\def \vgt {v_\textrm{gt}}
\def \vpred {v_\textrm{pred}}

\def \perceptual {L_\textrm{perc}}
\def \pix {L_\textrm{RGB}}
\def \sil {L_\textrm{sil}}
\def \restyle {L_\textrm{restyle}}
\def \vgg {L_\textrm{vgg}}
\def \ce {L_\textrm{ce}}
\def \contr {L_\textrm{contr}}

\def \predim {I}
\def \gtim {\hat{I}}

\def \predsil {S}
\def \gtsil {\hat{S}}

\newcommand{\loss}[1]{L_{\textrm{#1}}}
\newcommand{\weight}[1]{\lambda_{\textrm{#1}}}
\newcommand{\energy}[1]{\weight{#1}\loss{#1}}

\paragraph{Chamfer loss.} We sample 5000 vertices uniformly from the surface of each of the ground truth and predicted meshes, to get sampled vertices $\vpred$ and $\vgt$, and evaluate,

\begin{equation}
   \chamf = \sum_j{\min_i{\norm{v_{\textrm{gt},i}-v_{\textrm{pred},j}}_2}} + \sum_i{\min_j{\norm{v_{\textrm{gt},i}-v_{\textrm{pred},j}}_2}}
   \label{eq:chamf}
\end{equation}
\vspace{-6mm}
\paragraph{Smoothness loss.} We enforce a loss on the smoothness of the predicted mesh, which is a combination \change{ of a Laplacian smoothness \cite{nealen2006laplacian} metric, and an average edge length regularizer. }

\vspace{-2mm}
\paragraph{Texture loss.} We sample 1000 vertices $\vgt$ uniformly over the ground truth mesh surface, collect their colours $c_\textrm{gt}$, and enforce a loss encouraging the network to map $\vgt$ to $c_{\textrm{gt}}$,
\vspace{-3mm}
\begin{equation}
   \tex = \norm{c_\textrm{gt} - F(\vgt, \shapecode, \posecode, \texcode)_{\textrm{col}}}_2
\end{equation}

\vspace{-10mm}

\paragraph{Contrastive loss.} To encourage the latent pose space to meaningfully partition the data, we enforce a contrastive loss between selected pairs of pose latent codes $\posecode$. Given the latent vectors $z_{p,1}$ and $z_{p, 2}$, and corresponding `label' vectors $l_1, l_2$, we enforce the loss,
\vspace{-6mm}

\begin{equation}
\contr = yd^2 + (1-y) \max(M - d, 0)^2, \quad \textrm{where} \, d=\norm{z_{p,1} - z_{p, 2}}_2, y = l_1 \cdot l_2
\end{equation}
where $M=0.5$ is a selected margin distance to encourage between dissimilar pose vectors. Details of the formation of the labels $l$ can be found in the supplementary.

\vspace{-4mm}
\subsection{Training}
\vspace{-2mm}
Our full training loop consists of three stages: (1) registration, in which we optimise per-foot registration parameters to align our template mesh to the $Foot3D$ dataset; (2) network training, in which we optimise network weights $\theta$ and latent codes $\shapecode$, $\posecode$, and $\texcode$ in accordance with our training losses; and (3) latent refinement, where we fix network weights $\theta$ and fully optimise the latent codes to best match the data. In this loop, we minimise:
\vspace{-2mm}
\begin{equation}
\begin{matrix}
E_{\textrm{registration}}(r, t, s) = \energy{chamf}\\[5pt] 
E_\textrm{network training}(\theta, \shapecode, \posecode, \texcode) = \energy{chamf} + \energy{smooth} + \energy{tex}\\[5pt]
E_{\textrm{latent refinement}}(\shapecode, \posecode, \texcode) = \energy{chamf} + \energy{smooth} + \energy{tex}
\end{matrix}
\end{equation}
We select $\weight{chamf}=10^4, \weight{smooth}=10^3, \weight{tex}=1$, and train our network and latent parameters with an Adam optimizer \cite{kingma2014adam}, with a learning rate of $5 \times 10^{-5}$.

\vspace{-6mm}

\change{
\paragraph{Levels of supervision.} We are able to train our model with three levels of supervision, to provide better pose disentanglement and control: (i) Fully unsupervised, training only from mesh data; (ii) Using foot identity as a weak label, fixing shape between feet of the same identity for disentanglement of shape and pose; (iii) Adding pose descriptions as an additional weak label to allow for a more interpretable pose space via our contrastive loss.
}

\vspace{-7mm}
\section{Experiments}
\vspace{-2mm}
\paragraph{Foot3D dataset.} Our training set contains \ntpose{} feet in which the users were instructed to hold a neutral, `T-pose' position, and \narticulated{} feet total, including articulated feet. We use a further 8 feet for validation, of which 4 are `T-pose'.
\vspace{-5mm}
\paragraph{Evaluation metrics. } We evaluate the shape accuracy of our models using three metrics: a mean chamfer distance between a predicted model and the ground truth (as in Equation \ref{eq:chamf}); a mean Euclidean distance between 6 hand-labelled keypoints on the model's template mesh and the ground truth meshes; and an Intersection over Union (IoU) metric on silhouttes produced from 50 rendered viewpoints sampled uniformly in an arc across the foot.

\pagebreak
\paragraph{Baseline.} In order to compare our method with typical PCA implementations, we generate a PCA model. To do this without dense correspondences, we first use a FoldingNet \cite{yang2018foldingnet} implementation fitted to all of our data to produce a set of 1600-vertex point clouds with learned weak correspondences, and fit a PCA model to this data.
\vspace{-4mm}

\section{Results}

\begin{figure}
   \centering
   \includegraphics[width=\textwidth]{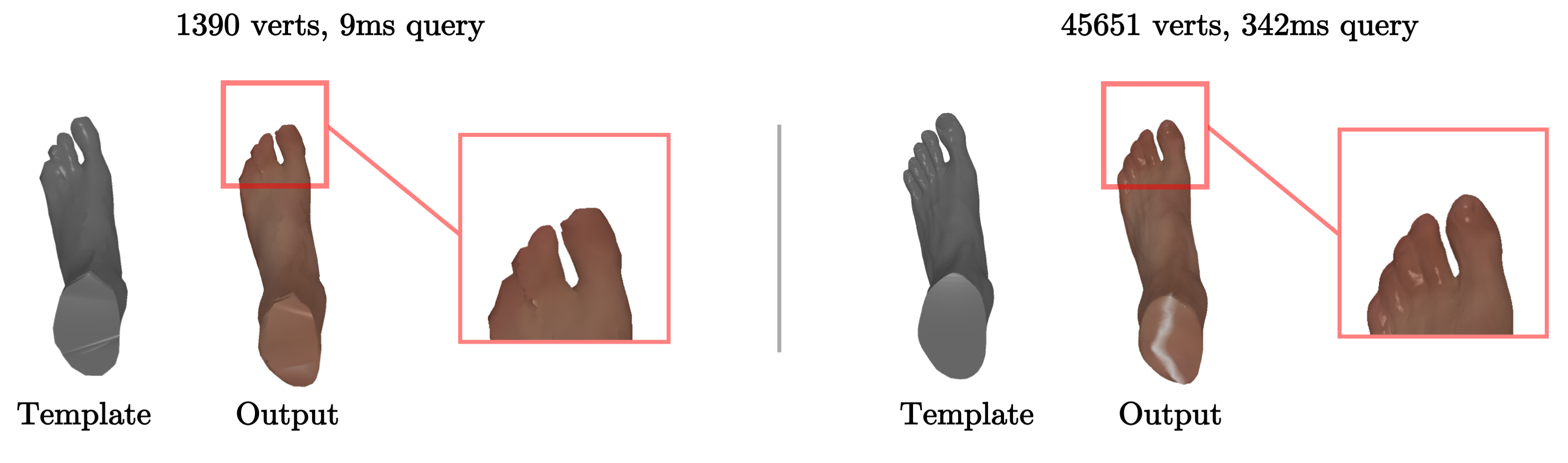}
   \vspace{-20pt}
   \caption{Our network can be sampled at different resolutions for lower-resolution applications (\eg{} mobile phones), or higher-resolution applications. Query times are shown on a Macbook M1 CPU.}
   \label{fig:sampling}
   \end{figure}

\begin{figure}[t!]
   \centering
   \includegraphics[width=\textwidth]{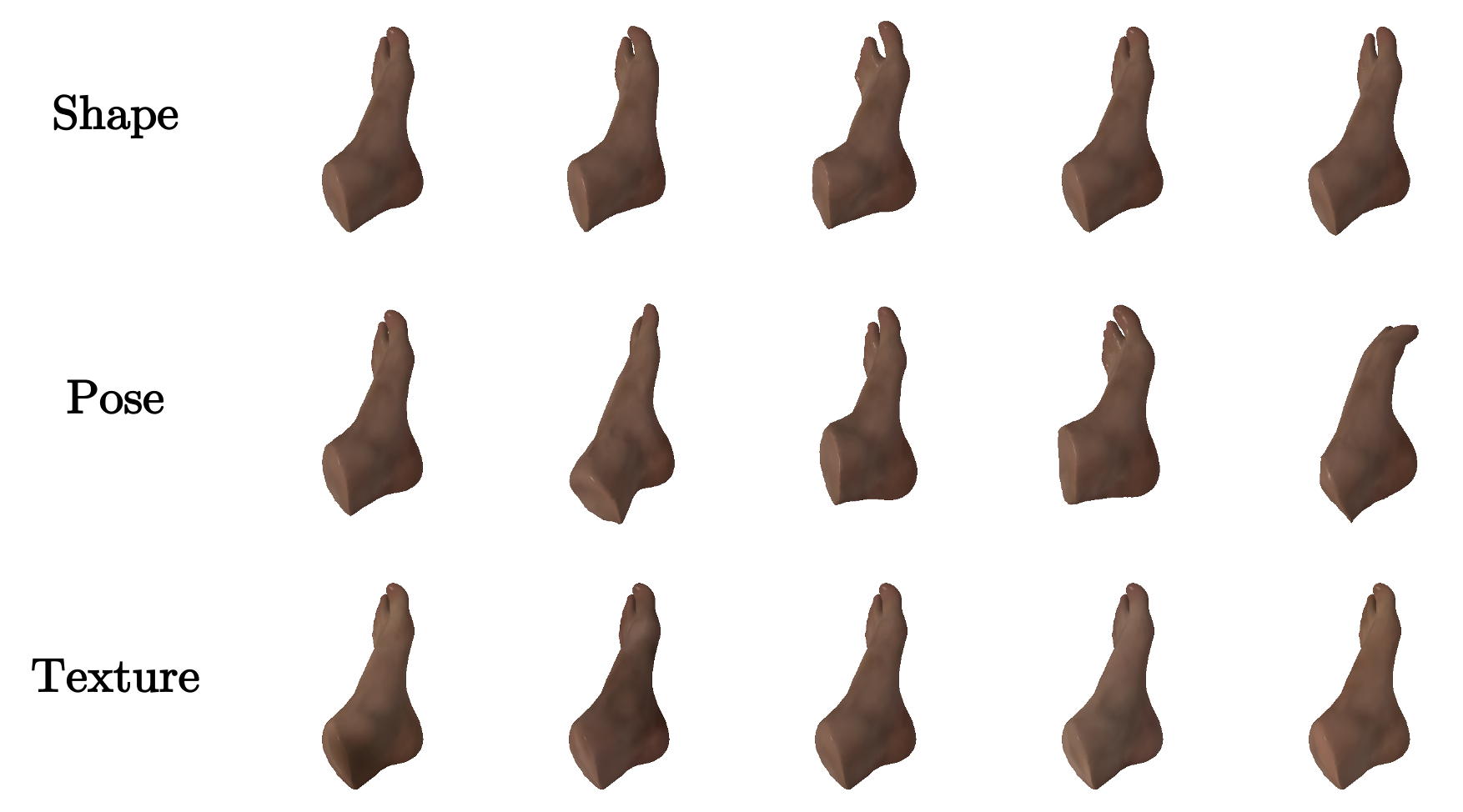}
   \vspace{-4mm}
   \centering
   \caption{Here we show 5 samples within the latent spaces of shape, pose and texture. Shape variations visible are toe lengths, separation and arch of the foot; pose variations visible are toe flexion and foot plantarflexion.}
   \label{fig:latent_space}
\end{figure}

\vspace{-1mm}

\paragraph{Multi-resolution.} As a coordinate based model, our neural surface deformation field can be sampled at as many 3D vertices as desired to produce an output mesh. As visualised in Figure \ref{fig:sampling}, we leverage this to sample at high resolutions for precise applications, or at lower resolutions for memory and CPU-critical applications, such as mobile phones. This requires template mesh to be reproduced at different resolutions, possible through off-the-shelf mesh simplification and subdivision tools \cite{Meshlab}.

\begin{figure}[t!]
   \centering
   \includegraphics[width=0.6\textwidth]{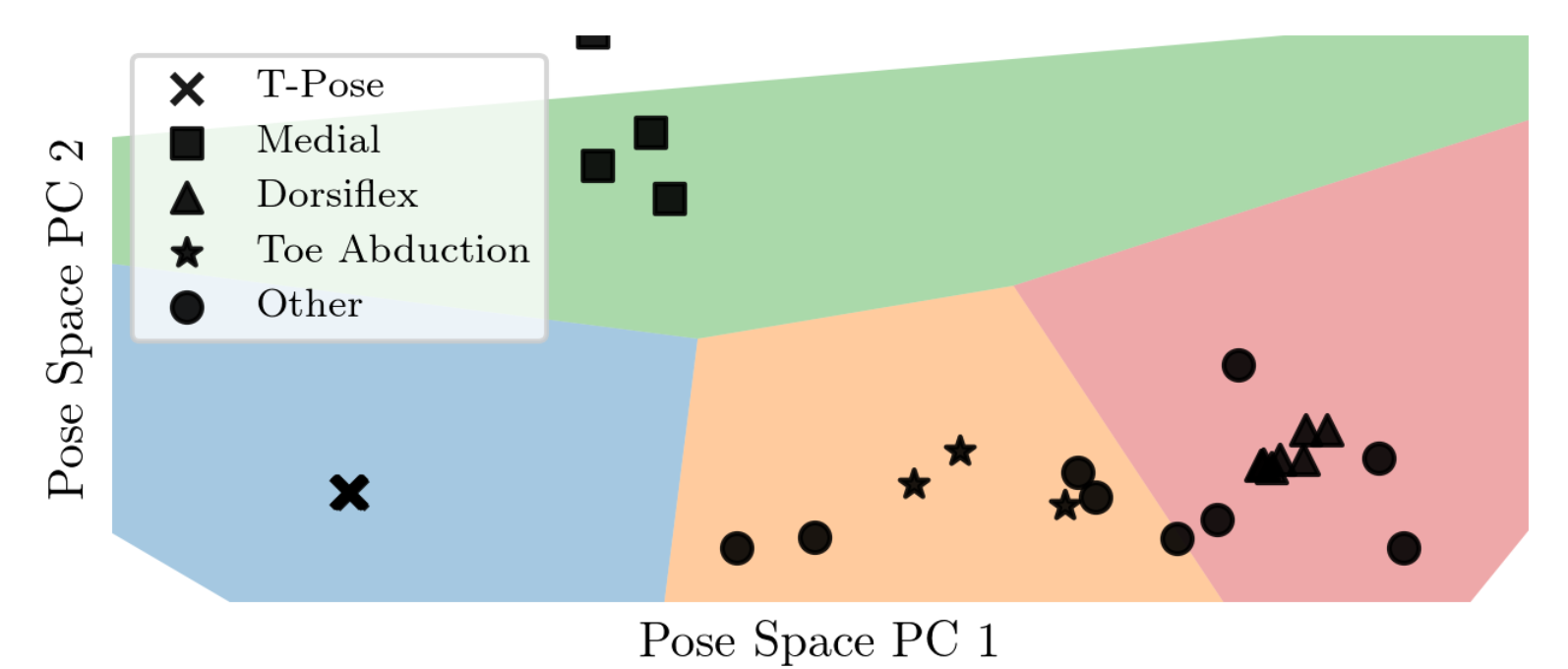}
   \vspace{-1mm}
   \caption{We visualise the FIND pose space on all training data in its two principal directions. We apply k-means clustering with 4 clusters to show partitioning of the space. We identify feet with poses in one of four pose categories to showcase the effective pose space partitioning due to our contrastive loss.}
   \label{fig:pose_clusters}
   \vspace{-1mm}

\end{figure}

\paragraph{Latent spaces.} Visualised learned shape, pose, and texture spaces of \MODEL{} are shown in Figure \ref{fig:latent_space}. Observe that our model has learned to disentangle shape and pose with minimal supervision using the simple constraint of fixing shape and texture codes between scans of feet belonging to the same person. Furthermore, we show in Figure \ref{fig:pose_clusters} that our contrastive loss ensures pose space is suitably segmented according to pose type.

\newcommand{\qualcol}[1]{\centering\hbox{\includegraphics[height=150pt]{images/qual/cr_columns/#1.png}}}
\newcolumntype{P}[1]{>{\centering\arraybackslash}p{#1}}
\begin{figure}[t]
    \centering
   \begin{tabular}{P{0.3\textwidth}|P{0.3\textwidth}|P{0.3\textwidth}}

   \begin{tabular}{*{3}{P{0.04\textwidth}}}
   Top & Side & Zoom \\
   \qualcol{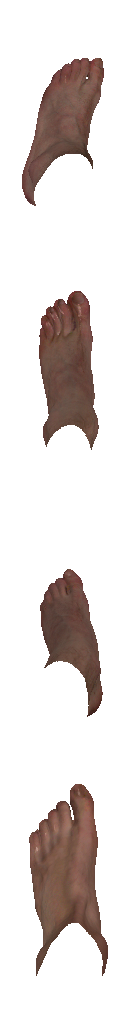} & \qualcol{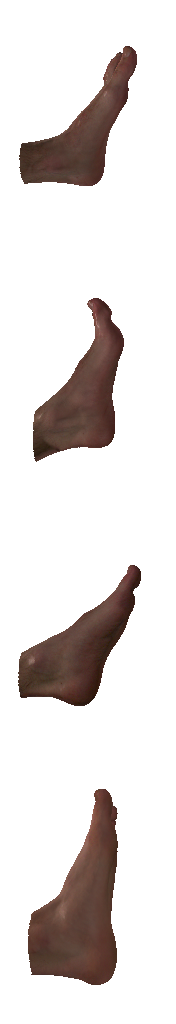} & \qualcol{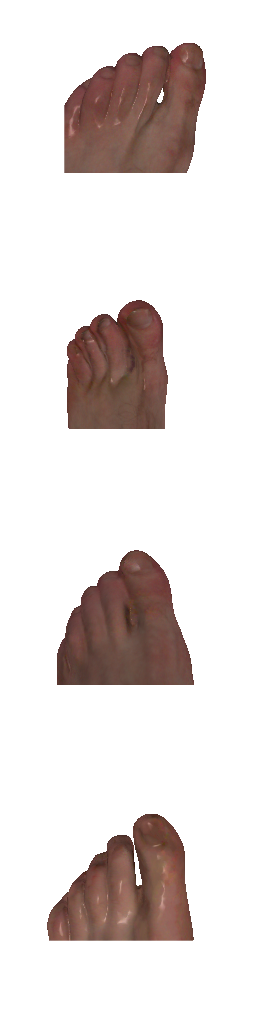}
   \end{tabular}
    &
   \begin{tabular}{*{3}{P{0.04\textwidth}}}
   Top & Side & Zoom \\
   \qualcol{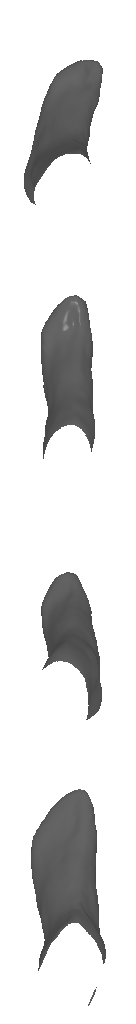} & \qualcol{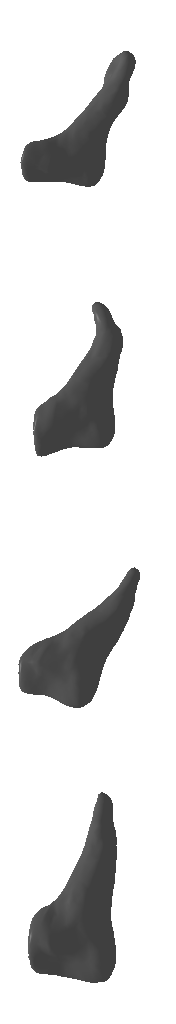} & \qualcol{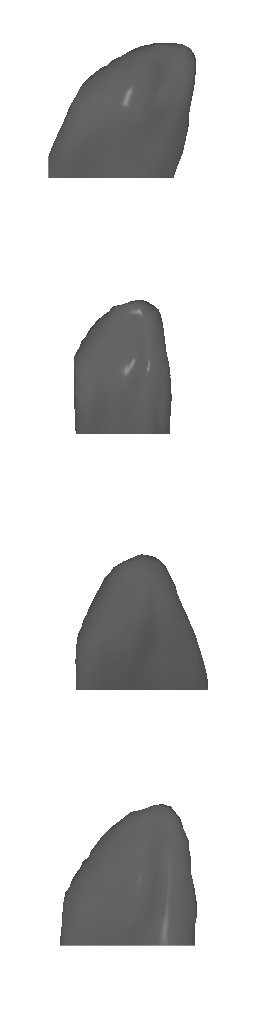}
   \end{tabular}
      &
   \begin{tabular}{*{3}{P{0.04\textwidth}}}
      Top & Side & Zoom \\
      \qualcol{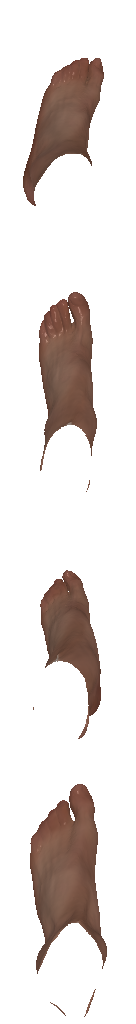} & \qualcol{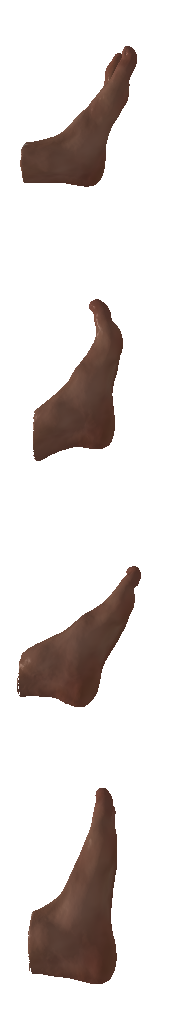} & \qualcol{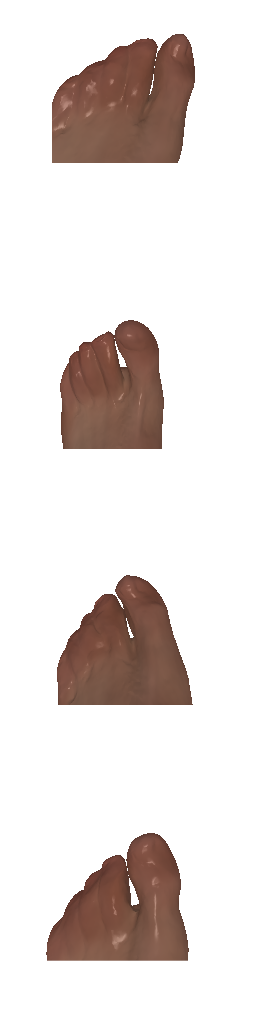}
      \end{tabular}
   \\[-3mm]
   (a) GT & (b) PCA & (c) Ours 
   \end{tabular}
   \vspace{-1mm}
   \caption{Qualitative results of our 3D fits to 8 validation feet, rendered from 3 views each.}
   \label{fig:qual3d}
\end{figure}

\begin{table}[t!]
   \footnotesize
  \centering
  \begin{tabular}{ccccc}
  \toprule
  Model & Trained on & Chamfer, mm $\downarrow$ & Keypoint, mm $\downarrow$ & IoU $\uparrow$\\
  \midrule
  PCA & T-Pose only & 4.5 & 15.0 & 0.892 \\
  PCA & All & 4.3 & 15.7 & 0.892 \\
  Ours & T-pose only & 3.8 & 6.4 & 0.923 \\
  Ours & All & \textbf{3.5} & \textbf{5.9} & \textbf{0.931} \\

  \bottomrule
  \end{tabular}
  \vspace{1mm}
  \caption{Comparison of \MODEL{} and a PCA baseline on our 8 validation feet.}
  \vspace{-5mm}
  \label{tab:3D_results}
\end{table}

\vspace{-4mm}
\paragraph{3D evaluation.} We compare our model to the PCA baseline. For both, we train 2 models: one on the \ntpose{} `T-pose' feet in the training dataset, and one on all \narticulated{} feet. Our `T-pose` trained model is restricted from learning a pose space. Table \ref{tab:3D_results} shows a significant quantitative improvement over the baseline, and Figure \ref{fig:qual3d} shows qualitative results. Notice our model does a much better job of toe separation.

\begin{table}

   \centering
   \small{
   \begin{tabular}{c|cc|cc}
   \toprule
   & \multicolumn{2}{c}{2 view} & \multicolumn{2}{c}{5 view} \\
   Optimisation loss & Chamfer, mm $\downarrow$ & Keypoint, mm $\downarrow$ & Chamfer, mm $\downarrow$ & Keypoint, mm $\downarrow$\\
   \midrule
   Sil & 9.0 & 14.4 & 4.1 & 7.7\\
   Sil + VGG & 8.9 & 13.1 & \textbf{4.0} & 7.3\\
   Sil + CE Loss & \textbf{6.8} & \textbf{10.3} & \textbf{4.0} & \textbf{6.4} \\
   \bottomrule
   \end{tabular}
} 
   \vspace{1mm}
   \caption{Quantitative results on inference pipeline on our validation feet.}
   \label{tab:inference_results}

\end{table}

\begin{table}[t!]
\small
   \centering
   \begin{tabular}{lccc}
   \toprule
   Model & Chamfer, mm $\downarrow$ & Keypoint, mm $\downarrow$ & IoU $\uparrow$ \\
   \midrule
   Full model & \textbf{3.5} & \textbf{5.9} & \textbf{0.931} \\
   $-$ Pose vector & 3.6 & 6.2 & 0.926 \\
   $-$ Texture loss & 3.7 & 6.8 & 0.928 \\
   \bottomrule
   \end{tabular}
   \vspace{1mm}
   \caption{Ablation study.}
   \label{tab:ablation}
\end{table}

\paragraph{Inference evaluation.} We show in Table \ref{tab:inference_results} our results of optimising \MODEL{} to renderings of validation feet, using only 2D losses to fit the registration and latent parameters. In our experiments, we sample 2 and 5 viewpoints in an arc around the foot. We compare three training losses: silhouette only; silhouette + an off-the-shelf VGG-16 \cite{simonyan2014very} perceptual loss; and silhouette + our novel Cross Entropy (CE) \partloss{}. We show that our loss improves the quality of these reconstructions. We find that our loss has a more substantial improvement in the recovered shape when fewer views are available for optimisation - we see this as likely being due to the extra 3D information this loss has available about unseen viewpoints.

\subsection{Ablation study}
We investigate in Table \ref{tab:ablation} the effect on our full model of several features. \change{We observe that the shape and pose disentanglement improves the shape reconstruction quality. Additionally, the texture loss, which is only able to learn from vertex colours, is also responsible for some of the reconstruction quality. We predict that this might be due to the surface colours providing supervision for vertex correspondences.
}

\section{Conclusion}
We demonstrate that our neural implicit foot model, \MODEL, outperforms typical PCA implementations of statistical shape learning, providing a higher fidelity shape space, whilst also able to learn texture and pose with minimal supervision. Furthermore, we have shown that simple foot identity and pose labels can be used to disentangle pose and shape, and produce a meaningful, interpretable pose space. Finally, we have shown that our \partloss{} improves upon typical shape model inference methods by enforcing part consistency.

\section{Acknowledgements}
The authors acknowledge the collaboration and financial support of Trya Srl.

\pagebreak

\bibliography{cameraready}

\includepdf[pages=-]{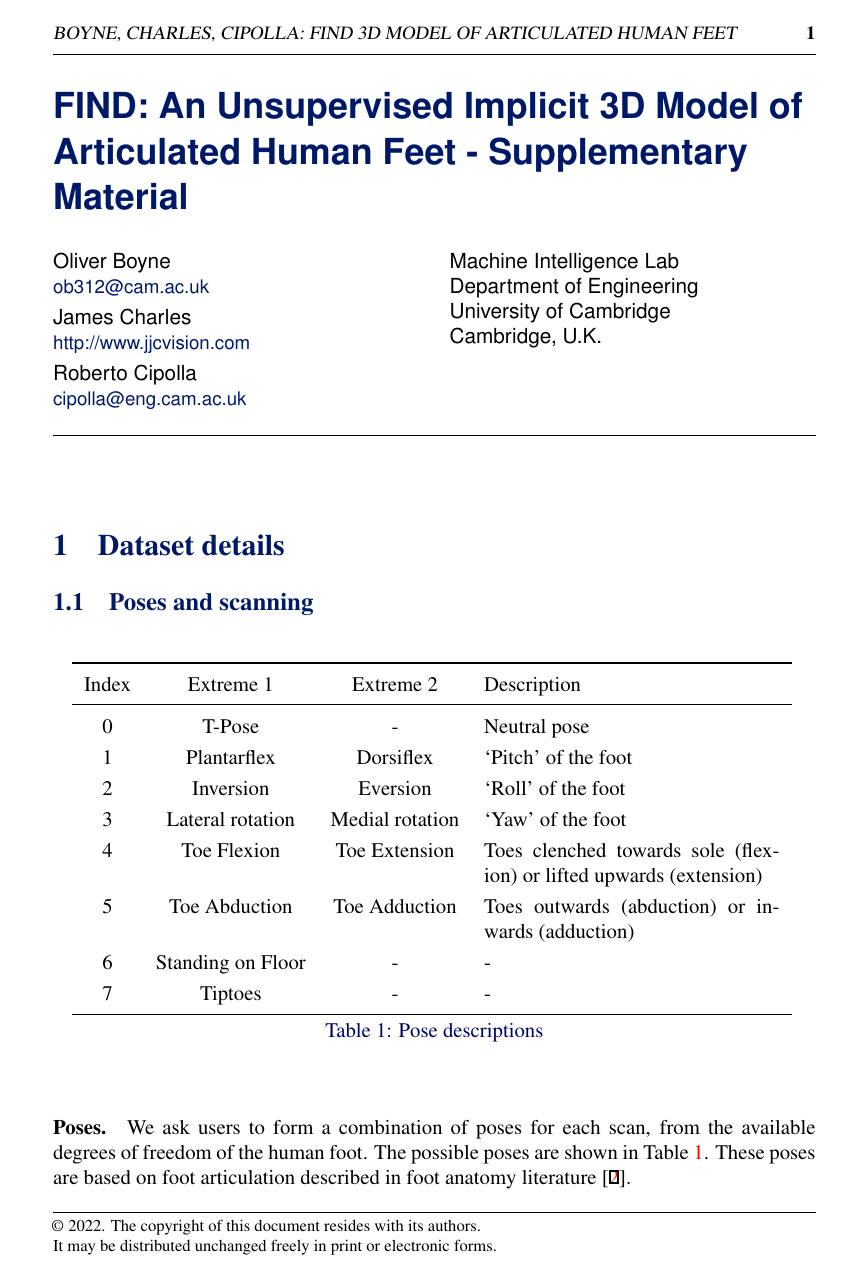}
\end{document}